\documentclass[10pt,twocolumn,letterpaper]{article}

\usepackage{cvpr}      

\usepackage[dvipsnames]{xcolor}

\definecolor{cvprblue}{rgb}{0.21,0.49,0.74}
\usepackage[pagebackref,breaklinks,colorlinks,citecolor=cvprblue]{hyperref}

\def\paperID{67} 
\def\confName{EarthVision}
\def\confYear{2024}

\title{Dense Air Pollution Estimation from Sparse in-situ Measurements and Satellite Data}

\author{Ruben Gonzalez Avilés\\
{\tt\small rubenoctavio.gonzalezaviles@student.unisg.ch}
\and
Linus Scheibenreif\\
{\tt\small linus.scheibenreif@unisg.ch}
\and
Damian Borth\\
{\tt\small damian.borth@unisg.ch}
\\\\
University of St. Gallen
}

\begin{document}
\maketitle
\begin{abstract}
This paper addresses the critical environmental challenge of estimating ambient Nitrogen Dioxide (NO\textsubscript{2}) concentrations, a key issue in public health and environmental policy. Existing methods for satellite-based air pollution estimation model the relationship between satellite and in-situ measurements at select point locations. While these approaches have advanced our ability to provide air quality estimations on a global scale, they come with inherent limitations. The most notable limitation is the computational intensity required for generating comprehensive estimates over extensive areas. Motivated by these limitations, this study introduces a novel dense estimation technique. Our approach seeks to balance the accuracy of high-resolution estimates with the practicality of computational constraints, thereby enabling efficient and scalable global environmental assessment. By utilizing a uniformly random offset sampling strategy, our method disperses the ground truth data pixel location evenly across a larger patch. At inference, the dense estimation method can then generate a grid of estimates in a single step, significantly reducing the computational resources required to provide estimates for larger areas. Notably, our approach also surpasses the results of existing point-wise methods by a significant margin of 9.45\%, achieving a Mean Absolute Error (MAE) of 4.98 \(\mu g/m^3\). This demonstrates both high accuracy and computational efficiency, highlighting the applicability of our method for global environmental assessment. Furthermore, we showcase the method's adaptability and robustness, by applying it to diverse geographic regions. Our method offers a viable solution to the computational challenges of large-scale environmental monitoring.
\end{abstract}

\begin{figure}[h]
    \centering
    \includegraphics[width=\columnwidth]{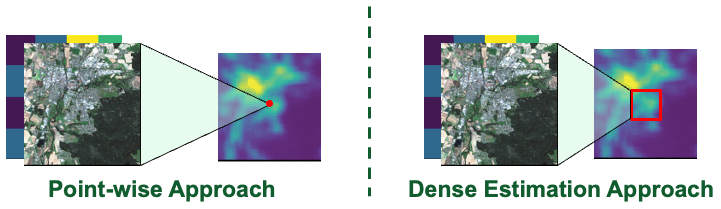}
    \caption{Comparison of point-wise and dense estimation approaches for NO\textsubscript{2} concentration estimation from satellite imagery. The point-wise approach (left) generates predictions at a single location, denoted by the red dot. The dense estimation approach (right) estimates NO\textsubscript{2} concentrations over a designated area, indicated by the red square, enhancing computational efficiency for regional-scale estimations.}
    \label{fig:point_vs_patch}
\end{figure}

\begin{figure*}[h]
    \centering
    \includegraphics[width=0.9\textwidth]{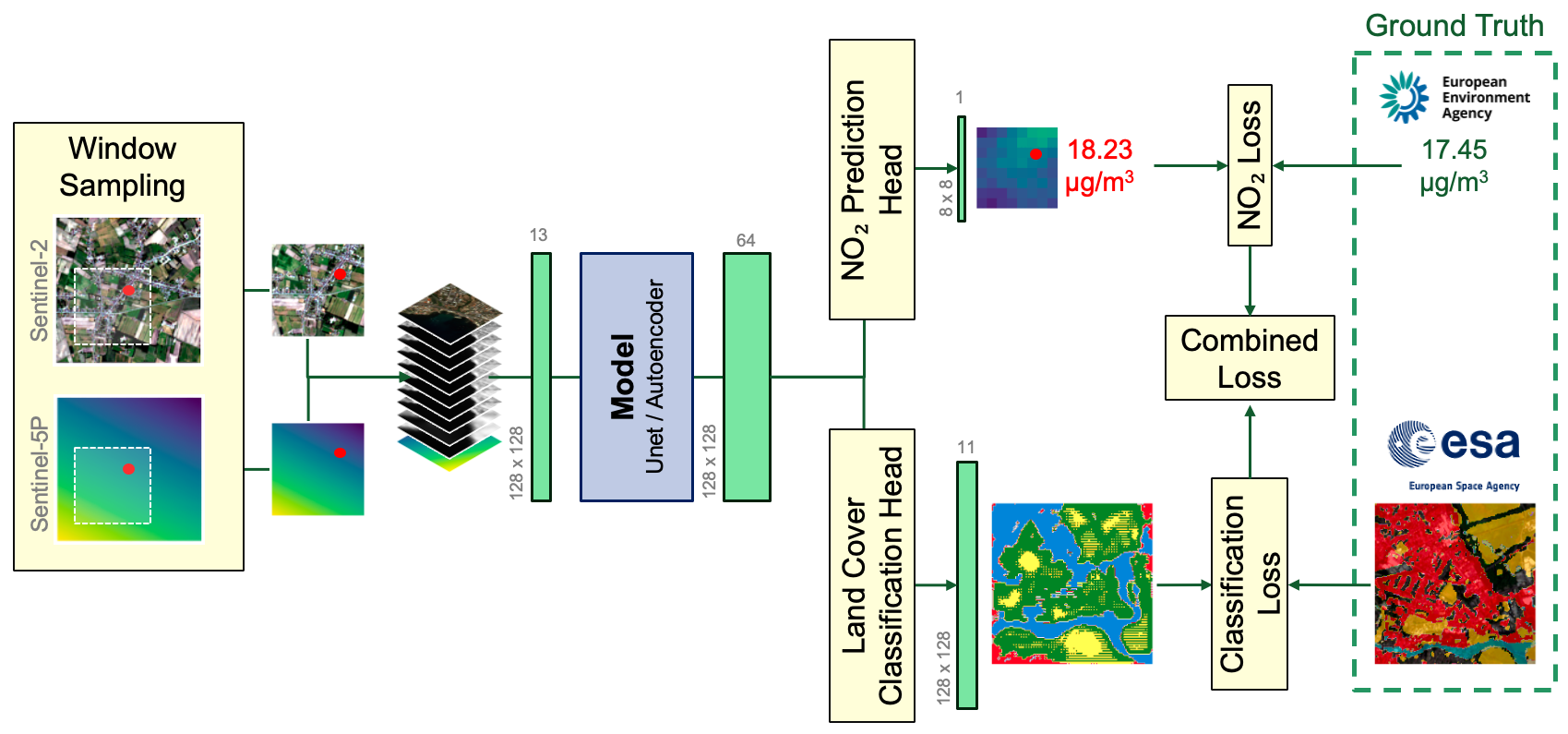}
    \caption{High-level overview showcasing the process of estimating NO\textsubscript{2} concentrations using a deep learning model. The model inputs are randomly sampled windows consisting of satellite data from Sentinel-2 and Sentinel-5P, which then pass through a convolutional neural network to extract relevant features. The NO\textsubscript{2} Regression Head produces a concentration estimate, which is subsequently aligned with ground truth data for validation. Concurrently, the Land Cover Classification Head processes the same features to classify land cover, aiding in the interpretation of NO\textsubscript{2} distribution. Losses from both heads are combined to optimize the model, with the ultimate goal of providing accurate, high-resolution estimations of ground-level NO\textsubscript{2}.}
    \label{fig:overview}
\end{figure*}
    
\section{Introduction}
\label{sec:intro}

Air quality degradation, particularly due to NO\textsubscript{2}, poses a significant risk to environmental sustainability and public health. This toxic and highly reactive gas, primarily emitted from the combustion of fossil fuels, is a major contributor to air pollution and precursor of greenhouse gases \cite{chen-2007, lammel1995greenhouse}. Moreover, NO\textsubscript{2} exposure is associated with a range of health issues. Various studies demonstrate the correlation between NO\textsubscript{2} levels and the prevalence of respiratory and cardiovascular diseases (see \cite{latza-2009} for a review). Therefore, the accurate assessment of NO\textsubscript{2} levels is crucial to safeguard public health and to support the development of effective policies aimed at mitigating its detrimental effects.

For these reasons, national public health and environmental agencies monitor NO\textsubscript{2} concentrations through networks of localized ground-based air quality stations. While ground-based instruments provide accurate and reliable data, they offer limited spatial coverage and incur high operational costs. These factors result in various notable data gaps, particularly in remote or less developed areas where monitoring stations are sparse \cite{lamsal-2008}.

To address these limitations, satellite-based remote sensing has emerged as a viable alternative for comprehensive environmental monitoring \cite{scheibenreif-2022, cooper-2020}. Satellites such as the European Space Agency's Sentinel-5P provide extensive global coverage, enabling the monitoring of atmospheric constituents, including NO\textsubscript{2}, on a much broader scale \cite{virghileanu-2020}. However, a significant challenge persists in effectively converting satellite observations into precise ground-level pollutant concentrations. This difficulty is particularly pronounced when estimating over extensive areas, due to the inherent spatial sparsity of available ground truth data, which are often confined to localized measurements.

This paper presents a novel method for ground-level NO\textsubscript{2} estimation, utilizing a dense estimation technique that leverages the capabilities of advanced deep learning models and the comprehensive data acquired from multi-spectral satellite imagery. We introduce a novel sampling technique to address the spatial sparsity of ground truth data, ensuring data continuity in areas without direct measurements. The dense estimation approach efficiently computes estimates for larger areas, significantly improving computational efficiency compared to individual point-wise estimations and enabling broader, more frequent assessment. This method marks a notable improvement in environmental monitoring, providing a scalable, efficient solution for global NO\textsubscript{2} estimation, which can in turn support enhanced environmental policy-making.

\section{Related Work}
\label{sec:related_work}

Local monitoring of NO\textsubscript{2} concentrations, primarily through ground-based stations, has been the traditional method for assessing air quality~\cite{guerreiro2014air}. Despite their precision, these localized stations face significant limitations, including their inability to provide comprehensive spatial coverage. This results in notable data blind spots, especially in remote and underdeveloped regions where setting up and maintaining such stations is logistically challenging and costly~\cite{verghese2022optimal}. Complementing this, the utilization of satellite data for environmental monitoring introduces its own set of complexities. While satellite imagery offers expansive geographical coverage, transforming these high-altitude observations into accurate, ground-level pollutant concentration estimates poses a significant challenge~\cite{li2011study}.

In the face of these limitations, researchers have increasingly turned to deep learning as a potential solution. Convolutional neural networks (CNNs), in particular, have demonstrated their efficacy in remote sensing applications, from land cover classification to air quality estimation \cite{zhu-2017}. These advancements in deep learning have opened up new possibilities for interpreting and utilizing satellite data more effectively for environmental monitoring purposes~\cite{tuia2023artificial,liao2020deep}.

Several recent methods have utilized deep learning approaches for air pollution estimation from satellite data at discrete geospatial locations, which we will henceforth refer to as the point-wise approach \cite{rowley2023predicting,scheibenreif-2022}. This methodology, as highlighted in studies by \citet{rowley2023predicting}, and \citet{scheibenreif-2022} excels in producing highly accurate pollution estimates at specific global locations. Despite its precision and significant improvement over traditional methods, the point-wise approach is constrained by its inherent focus on single spatial locations. When applied to expansive areas, the point-wise method necessitates repeated processing for each location, resulting in considerable computational demands for global-scale application.

Addressing the limitations inherent in the point-wise approach, our research introduces a more efficient alternative: the dense estimation approach. This novel method aims to maintain the accuracy of point-wise estimations while significantly reducing computational requirements. Instead of producing individual estimates, the dense estimation approach simultaneously generates a grid of estimates for a designated area, effectively decreasing the total number of iterations required for larger-scale NO\textsubscript{2} concentration assessments. The dense estimation approach thus adeptly strikes a balance between the need for accurate estimates with the practical limitations of computational resources.

\section{Approach}
\label{sec:approach}

Figure \ref{fig:overview} provides an overarching view of our approach to estimating ground-level NO\textsubscript{2} concentrations, combining advanced deep learning techniques with multispectral satellite imagery. In the following subsections, we will explore each component of this methodology in detail, explaining how they collectively contribute to an efficient and scalable environmental monitoring solution.

\begin{figure}[!t]
    \centering
    \includegraphics[width=\columnwidth]{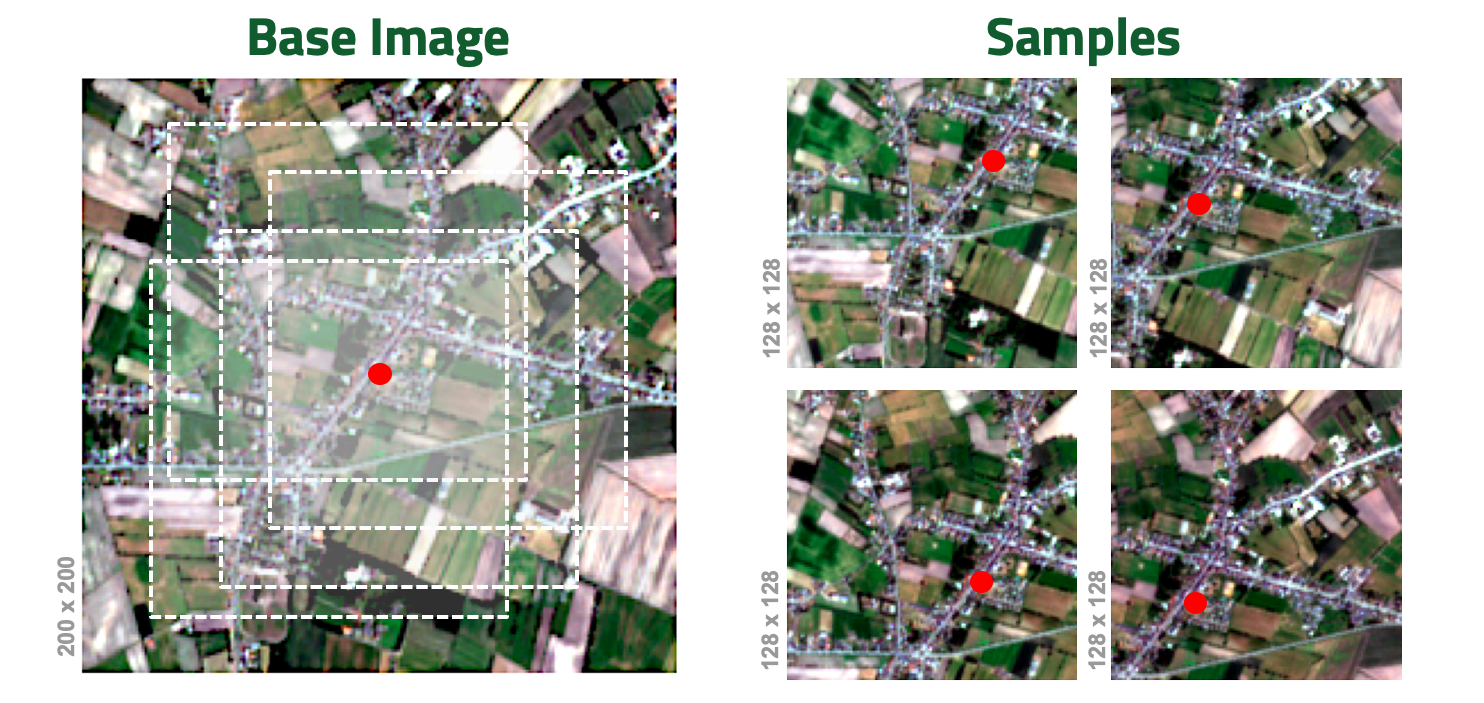}
    \caption{Illustration of the uniformly random offset sampling technique. The base image (left) shows the full 200 x 200 pixels area from which samples are taken. Individual samples (right) are 128 x 128 pixels, each randomly offset within the larger image. This sampling strategy diversifies the pixel location of the measurement point while retaining the original geospatial station coordinates.}
    \label{fig:sampling}
\end{figure}
\subsection{Handling Sparse Annotations}

\paragraph{The Challenge of Sparse Annotations} 
One of the primary challenges in estimating surface-level greenhouse gas concentrations from satellite imagery is the inherent sparsity of ground truth annotations. Typically, ground truth measurements are only available for a specific geographical location, which corresponds to the coordinates of the measurement station or device. Therefore, the ground truth only covers a single point within the expansive area captured by the satellite images. This sparsity presents a significant challenge in training models to accurately infer greenhouse gas concentrations across a broader area.

Another challenge within our dataset is the centralized positioning of ground truth annotations – each being at the center of the corresponding image. This unique setup poses a risk of the model developing a bias towards predicting accurate NO\textsubscript{2} levels only for the central pixel, while potentially yielding random estimates for the remaining pixels in the image. Such a scenario is problematic for our objective of estimating concentrations across an entire grid, rather than just at a single point. If unaddressed, this could result in a model that, while precise at the center, disperses random or less accurate estimations across the rest of the grid, undermining the goal of achieving comprehensive and uniform coverage in NO\textsubscript{2} concentration estimation.

\paragraph{Uniformly Random Offset Sampling Technique}
\label{subsec:randomsampling}

\begin{figure}[!t]
    \centering
    \includegraphics[width=\columnwidth]{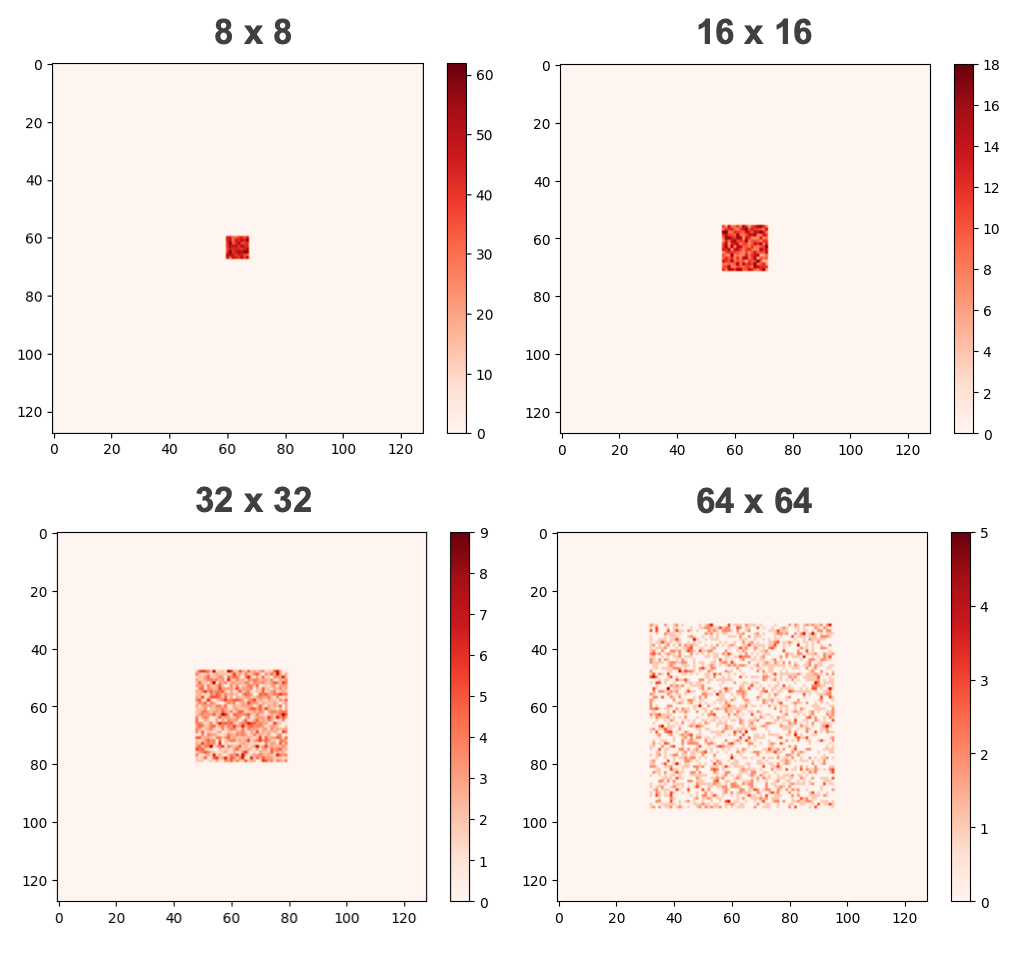}
    \caption{Distribution of ground-truth pixel coordinates of 2871 samples within the sampling window for varying prediction space sizes.}
    \label{fig:prediction_space}
\end{figure}

To address this issue, we employed a novel sampling technique that introduces variability into the spatial relationship between the satellite imagery and the ground truth measurements. This method involves sampling each image with a fixed-size window, ensuring that every sample maintains the same spatial dimensions (128 x 128 pixels). However, instead of centering this window on the measurement location, we implemented a uniformly random offset. This offset shifts the sampling window from the origin by an amount sampled from a uniformly random distribution.
Crucially, the range of this offset is constrained to ensure that the ground truth measurement locations are uniformly distributed over a designated area, henceforth referred to as the prediction area. This prediction area forms the scope of our model's prediction, ensuring that the model learns to estimate NO\textsubscript{2} concentrations across a broader and more representative area rather than being limited to a single central point.

Despite the random offset, each sample retains the information about the new pixel coordinates of the measurement location. This ensures that the ground truth remains accurately associated with the original geospatial location.

\subsection{Model Architecture}

To experiment with various outcomes, we developed two distinct neural network models, each exhibiting unique architectural capabilities. These models are designed to process and interpret satellite data, outputting a feature map that is further utilized by two specialized heads for dedicated tasks. 

\paragraph{Model I - UNet}
The first model is based on the UNet architecture, widely acclaimed for its effectiveness in image segmentation tasks. The UNet is characterized by its distinctive U-shaped design, incorporating a contracting path for capturing and downsampling features, complemented by an expansive path for reconstruction. In our adaptation of this architecture, padding has been purposefully incorporated, particularly within the contracting path. This modification ensures that the spatial dimensions of the input can be accurately reconstructed.

\paragraph{Model II - Autoencoder}
The second model is an Autoencoder architecture, derived from the UNet but with a significant modification: the exclusion of skip connections. This simplification aims to test the model’s capability in learning and reconstructing features without the direct transfer of information between corresponding layers of the encoder and decoder. Notably, the absence of skip connections in the Autoencoder avoids the introduction of fine-grained features that the UNet might otherwise capture. This characteristic is particularly relevant for estimating greenhouse gas concentrations since it is expected that the concentration levels do not exhibit strong variations from pixel to pixel. By eliminating these connections, the Autoencoder can potentially provide a more generalized and smoother representation of the NO\textsubscript{2} distribution, which could be more accurate for this specific task.

\paragraph{Input and Output Specifications}
Both the UNet and Autoencoder models process inputs comprising the 12 multispectral bands from Sentinel-2 data and an additional plane representing Sentinel-5P data. These inputs encompass the geographical area determined by the random sampling technique previously described in Section \ref{subsec:randomsampling}.

Each model, the UNet and the Autoencoder, generates a feature map with dimensions of 128 x 128 x 64 in its output. Both models benefit from the expanded depth of 64 channels in their outputs, which allows for a richer representation of the processed satellite imagery.

\subsection{Dual-Task Heads}

Upon processing the input data, the outputs of these models are then fed into two distinct task heads, each serving a specific purpose:

\begin{figure}[!h]
    \centering
    \includegraphics[width=\columnwidth]{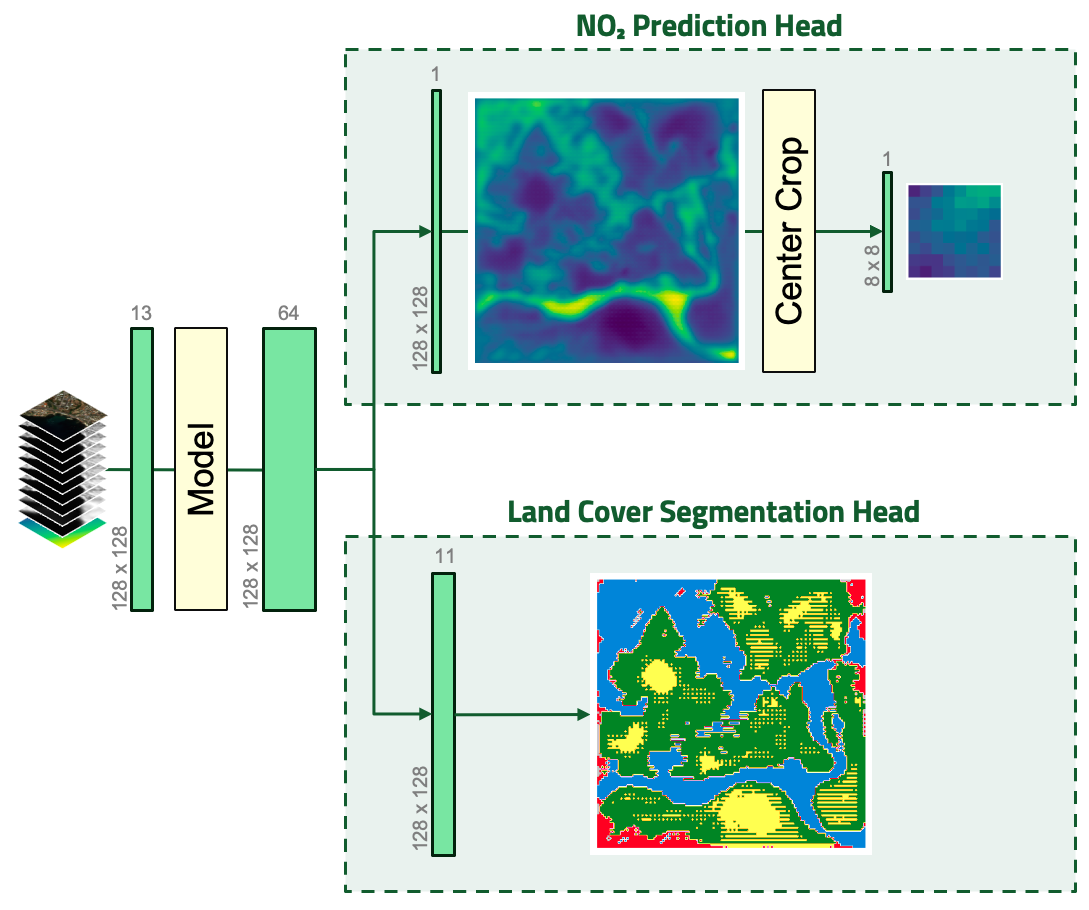}
    \caption{Illustration showcasing the dual-task approach where one head is dedicated to the NO\textsubscript{2} estimation, and the other head focuses on land cover classification.}
    \label{fig:dual_head}
\end{figure}

\paragraph{NO\textsubscript{2} Prediction Head}
This head utilizes the model output to estimate the NO\textsubscript{2} concentration at the ground level. It focuses on producing a 2-dimensional regression plane, reflecting the spatial distribution of NO\textsubscript{2} concentrations across the sampled area. The output from this head is then further center-cropped to match the dimensions of the prediction area. This prediction area is determined by the distribution of measurement coordinates, a result of the random sampling technique as detailed in Section \ref{subsec:randomsampling}. Such cropping ensures that the model's output comprehensively encompasses the entire area where the ground truth data coordinates are distributed in, thereby aligning the predictions accurately with the spatial scope of the ground truth.

After the center-cropping process, the output of the head consists of a plane with the spatial dimensions of the prediction area and a depth of 1. Thus, the final output effectively covers the entire area of interest, providing a detailed map of surface-level NO\textsubscript{2} concentration predictions.

\paragraph{Land Cover Segmentation Head}
The second head is responsible for classifying the land cover within the geographical area captured by the input sample. While the land cover segmentation results are utilized exclusively during the training phase and not in the inference process, they potentially enhance the model's learning. This approach is rooted in the hypothesis that NO\textsubscript{2} concentrations may be closely related to the type of land cover. By training the model to recognize these land cover categories, it is hypothesized that the model might develop a more nuanced understanding of potential NO\textsubscript{2} distribution patterns. This understanding, based on the underlying environmental context, could potentially improve the overall accuracy and effectiveness of the estimation. Utilizing the feature map generated by the models, this head outputs a 128 x 128 x 11-dimensional representation, where each pixel is categorized into one of 11 distinct land cover classes, thus providing a complete prediction of the land cover diversity within the sampled area.


\subsection{Loss Function Formulation}

A critical component of the model's training process is the choice of the loss function, which guides the optimization of the models. For our research, a combined loss function was implemented, tailored to accommodate the dual objectives of the NO\textsubscript{2} concentration estimation and land cover segmentation.
The combined loss function comprises two primary components: the Squared Error Loss for the NO\textsubscript{2} prediction task, and the Cross Entropy Loss for the land cover segmentation task. Each component is designed to optimize the model's performance in its respective task.

\paragraph{Squared Error Loss (NO\textsubscript{2} Regression)}
For the NO\textsubscript{2} concentration estimation, the Squared Error Loss (SEL) is used, mathematically represented as:
\begin{equation}
    \text{SEL}(y, \hat{y}_{i,j}) = (y - \hat{y}_{i,j})^2
    \label{eq:SEL}
\end{equation}
where \( y \) is the ground truth NO\textsubscript{2} concentration and \( \hat{y}_{i,j} \) is the model's predicted concentration at the pixel location \( (i, j) \). This loss function penalizes the discrepancies between the predicted and actual NO\textsubscript{2} levels, with a larger penalty for more significant errors.

\paragraph{Cross Entropy Loss (Land Cover Segmentation)}

The Cross Entropy Loss (CEL) is employed for the land cover segmentation task, defined as:
\begin{equation}
    \text{CEL}(\hat{c}, c) = -\sum_{i=1}^{C} w_i \cdot {1}(c = i) \cdot \log\left(\frac{\exp(\hat{c}_i)}{\sum_{j=1}^{C} \exp(\hat{c}_j)}\right)
    \label{eq:CEL}
\end{equation}
In this formula, \( \hat{c} \) represents the predicted class probabilities, \( c \) is the true class, \( C \) is the number of classes, \( w_i \) are the class weights, and \( {1}(c = i) \) is an indicator function that is 1 if \( c = i \) and 0 otherwise. This loss function is a weighted variant of the standard cross entropy loss, designed to account for the class imbalance present in the ground truth data. The weighting \( w_i \) for each class compensates for the prevalence of certain land cover types, ensuring a more balanced learning process and improving the model's ability to classify each pixel accurately into the correct land cover category.

\paragraph{Combined Loss Function} 
The final loss function used in the model training is a weighted sum of these two components:
\begin{equation}
    \text{L}(y, \hat{y}_{i,j}, c, \hat{c}, \lambda) = \text{SEL}(y, \hat{y}_{i,j}) + \lambda \times \text{CEL}(\hat{c}, c)
    \label{eq:L}
\end{equation}
where \( \lambda \) is a regularization parameter that balances the contribution of each loss component. This combined loss function allows simultaneous optimization for both NO\textsubscript{2} concentration estimation and land cover segmentation, ensuring that the model is effectively trained to perform both tasks.


\section{Experimental Results}
\label{sec:results}

\subsection{Dataset}

\paragraph{Ground-Level NO\textsubscript{2} Measurements}
The core of the training data consists of surface-level NO\textsubscript{2} concentration measurements, expressed in micrograms per cubic meter (\(\mu g/m^3\)). These measurements were sourced from 3087 unique locations, providing a diverse and geographically varied dataset. The data consists of measurements taken at an hourly frequency, which have been averaged over a time frame spanning from 2018 to 2020 \cite{scheibenreif_2022_5764262}.

\paragraph{Satellite Imagery}
The ground truth measurements were complemented with high-resolution satellite imagery from two distinct satellite missions:

\paragraph{Sentinel-2 Data}
This includes 12 multispectral bands with a resolution of 10 meters, offering detailed visual and spectral information about the Earth's surface. Each image segment corresponds to a 200 x 200 pixel area (approximately 2 x 2 km), cropped around the measurement location. To ensure data quality, the data was pre-selected to only include images with minimal cloud coverage and artefacts \cite{scheibenreif_2022_5764262}.

\paragraph{Sentinel-5P Data}
Sentinel-5P data provides insights into trace gases and aerosols in the atmosphere, offering a spatial resolution of 5 x 3.5 km\(^2\). To facilitate compatibility with Sentinel-2 imagery, this data has been linearly interpolated to a 10-meter resolution \cite{scheibenreif_2022_5764262}. 


\paragraph{WorldCover Data}
Additionally, the WorldCover dataset \cite{zanaga_2022_7254221} was utilized for land cover classification information. This dataset provides a classification of each pixel into one of 11 classes and was aligned to cover the same area as the Sentinel-2 data. The integration of WorldCover data supports the model's understanding of land cover variations and their potential impact on NO\textsubscript{2} distribution.

\paragraph{Data Preprocessing and Augmentation}
Prior to training, all data underwent preprocessing to ensure consistency and compatibility. This included normalization of the satellite imagery and ground truth measurements. Furthermore, the uniformly random distributed offset sampling technique, as detailed in section \ref{subsec:randomsampling}, was applied to augment the dataset and mitigate the issue of sparse annotations.

\subsection{Training Methodology}

\paragraph{Dataset Size and Split}
The full dataset comprises a total of 3087 samples, each representing a unique combination of ground-level NO\textsubscript{2} measurements and corresponding satellite data collected by Scheibenreif et al. (2022) \cite{scheibenreif_2022_5764262}. However, since the land cover ground truth data was not available for every measurement location, the effective dataset resulted in 2871 complete samples. The division of this dataset was strategically executed as follows:

\begin{itemize}
    \item Training set: 70\% (2009 samples)
    \item Validation set: 15\% (431 samples)
    \item Test set: 15\% (431 samples)
\end{itemize}

This split ensures a substantial training dataset for model learning, while providing adequate and equal-sized data for validation and testing.

\paragraph{Hyperparameter Tuning and Evaluation Metrics}
Hyperparameter tuning was conducted using the validation dataset, focusing on optimizing parameters such as learning rate, batch size, prediction space and the regularization parameter \(\lambda\) in the loss function. The primary evaluation metric was the Mean Absolute Error (MAE) for the NO\textsubscript{2} concentration estimation on the validation set.

\paragraph{Baseline and Training Procedure}
The baseline for this study was established based on previous research conducted by Scheibenreif et al. (2022) \cite{scheibenreif-2022}. It is important to note that the aim was not to surpass this baseline but to explore the feasibility and effectiveness of the proposed approach, considering that the prediction area in this study is larger, thus posing a more challenging task.

The training procedure involved assessing both models using two distinct loss functions — either a combined loss approach \eqref{eq:L} or solely the squared error loss \eqref{eq:SEL} computed on the resulting NO\textsubscript{2} predictions. The models were trained using a GPU-accelerated environment to ensure efficient learning. Regular checkpoints were saved for model performance evaluation and to facilitate the selection of the best-performing model based on validation data.

\subsection{Evaluation}

In this subsection, we conduct a comprehensive comparative analysis of the point-wise and dense estimation approaches, incorporating both qualitative and quantitative evaluations.

\begin{figure}[h]
    \centering
    \includegraphics[width=\columnwidth]{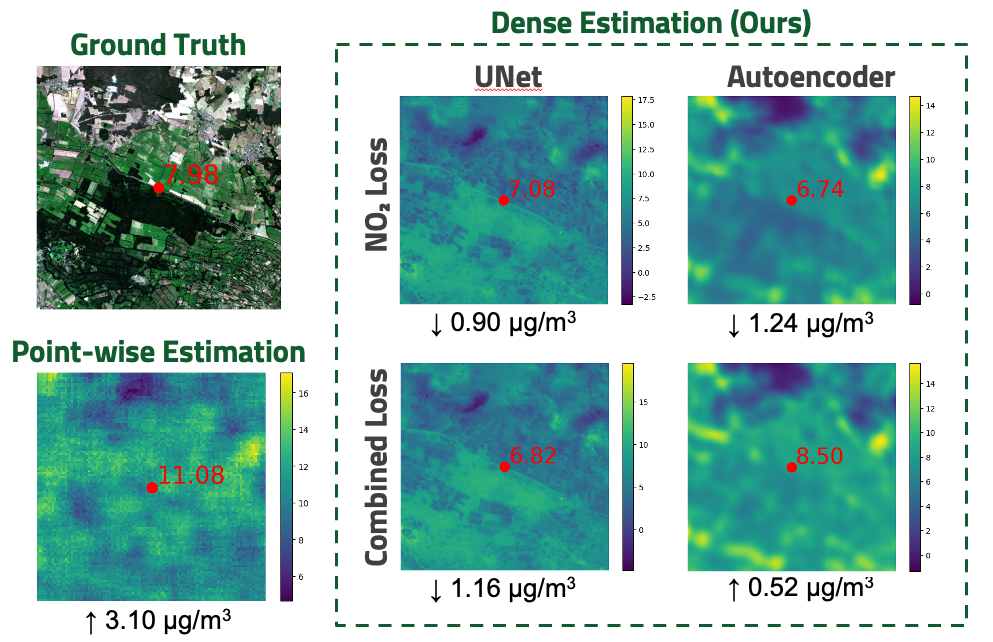}
    \caption{Comparative visualization of NO\textsubscript{2} estimation methods: the ground truth with a marked measurement point, point-wise estimation showing per-pixel predictions, and dense estimation results by UNet and Autoencoder models with different loss objectives. Deviations from the ground truth are indicated below each estimate.}
    \label{fig:model_comparison_results}
\end{figure}

\paragraph{Qualitative Analysis}
The qualitative analysis focuses on the characteristics and visual aspects of the results generated by each approach, as illustrated in Figure \ref{fig:model_comparison_results} which showcases the different outcomes based on the same input area. The input area for this analysis comprises data with spatial dimensions of 1200 x 1200 pixels, corresponding to an area of 1.2km x 1.2km. For the dense estimation approach, predictions were achieved by iteratively creating prediction patches of 8x8 pixels and concatenating these to form the final output. In contrast, the point-wise results were generated by iteratively estimating several points within the input area and employing linear interpolation to achieve a cohesive prediction across the entire area \cite{scheibenreif-2022}.

\paragraph{UNet Model}
The UNet model, employed in the dense estimation approach, exhibits a tendency to generate a more fine-grained distribution of NO\textsubscript{2} concentrations. While this high level of detail can be advantageous in certain applications, it may not be ideally suited for NO\textsubscript{2} prediction. This finer granularity could potentially introduce localized variations, which might not be representative of the broader environmental context of NO\textsubscript{2} distribution.

\paragraph{Autoencoder Architecture}
In contrast, the Autoencoder architecture demonstrates a distribution pattern that aligns more closely with the results from the point-wise approach. This similarity suggests that the Autoencoder is capable of capturing the essential features relevant to NO\textsubscript{2} concentration without the excessive detail that might complicate the interpretation or application of the data. Its performance indicates a balance between detail and contextual accuracy, potentially making it more suitable for reliable greenhouse gas predictions.

\begin{figure}[h]
    \centering
    \includegraphics[width=\columnwidth]{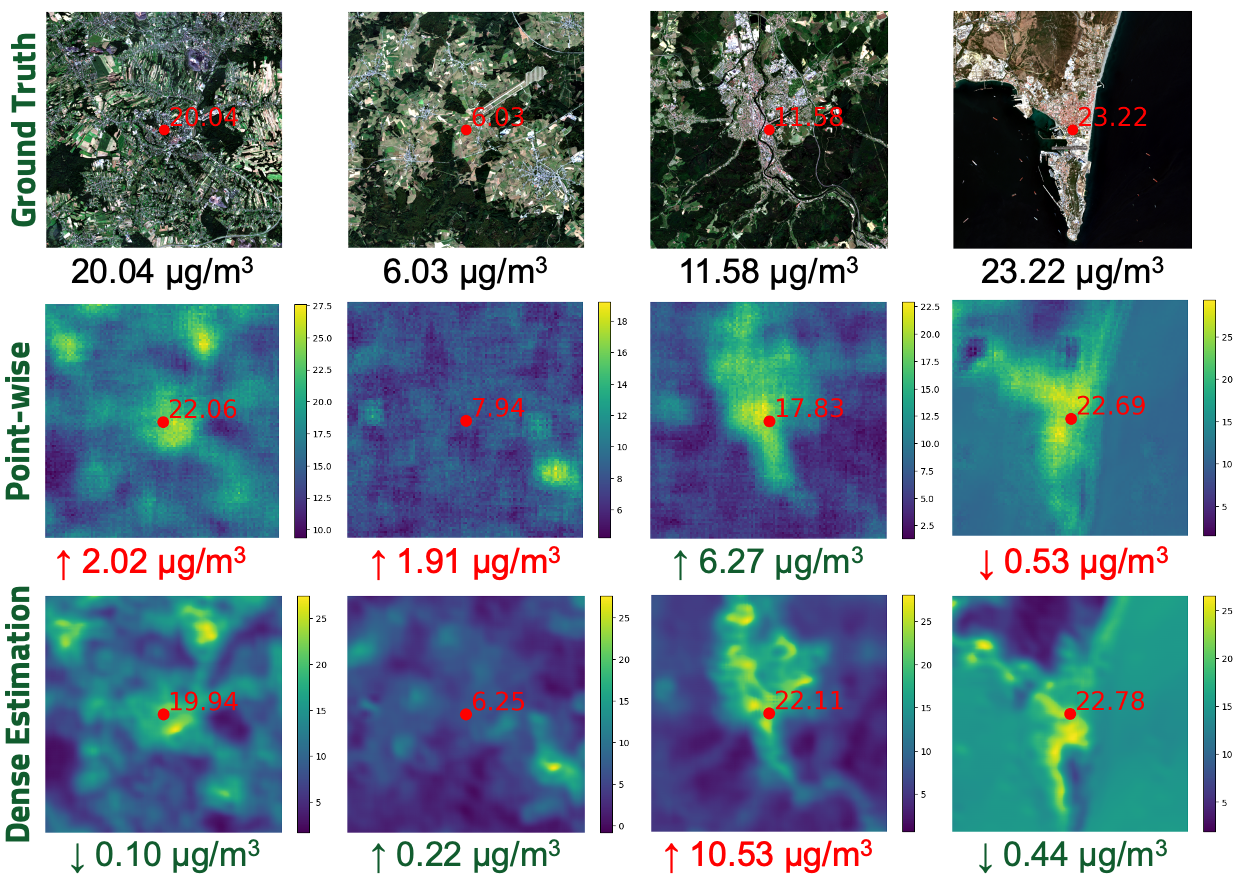}
    \caption{Visualization of ground truth NO\textsubscript{2} levels and corresponding point-wise and dense estimation results using the Autoencoder model with NO\textsubscript{2} loss, showing close approximation to actual concentrations and similar distributions as the point-wise estimates.}
    \label{fig:point_vs_patch_results}
\end{figure}

\textbf{Quantitative Analysis}
In the quantitative analysis, we examine the performance metrics of both approaches, including MAE, R2-Score, and Mean Squared Error (MSE). These evaluation results are depicted in Table \ref{tab:point_vs_patch} and enable a comprehensive understanding of each model's accuracy and predictive capabilities.

\begin{table}[!ht]
  \centering
  \begin{tabular}{@{}l c c c c}
    \toprule
\textbf{Model} & \textbf{Loss} & \textbf{R2 ($\uparrow$)} & \textbf{MSE ($\downarrow$)} & \textbf{MAE ($\downarrow$)} \\
    \midrule
UNet (Ours) & Combined & 0.49 & 47.25 & 5.06 \\ 
UNet (Ours) & NO\textsubscript{2} & 0.49 & \textbf{46.83} & 5.03 \\ 
\textbf{AE (Ours}) & Combined & 0.47 & 50.94 & \textbf{4.98} \\ 
AE (Ours) & NO\textsubscript{2} & 0.46 & 49.55 & 4.99 \\ 
\multicolumn{2}{@{}l}{Point-wise \cite{scheibenreif-2022}} & 0.55 & 61.25 & 5.65 \\ 
\multicolumn{2}{@{}l}{Point-wise (Pre-trained) \cite{scheibenreif-2022}} & \textbf{0.57} & 58.47 & 5.50 \\ 

    \bottomrule
  \end{tabular}
  \caption{Quantitative Results: Point-wise vs Dense Estimation.}
  \label{tab:point_vs_patch}
\end{table}

The quantitative analysis reveals that our dense estimation approach not only matches but surpasses the accuracy of the point-wise approach, achieving significantly lower values in both the MAE and MSE metrics. This is particularly noteworthy considering our model's operational conditions, which included a reduced training set and an expanded test set (10\% for point-wise \cite{scheibenreif-2022} and 15\% for our dense estimation approach), further underscoring the effectiveness and robustness of our method.

\paragraph{Summary}

In summary, the qualitative and quantitative analyses highlight the strengths and advantages of the dense estimation approach over the point-wise method. While the point-wise approach already excelled in accuracy, our dense estimation method, particularly with the Autoencoder architecture, now not only equals but surpasses it in both visual and quantitative metrics. This advancement is achieved with significantly lower computational costs. Our dense estimation method achieves this by processing patches of 64 pixels (8x8) collectively, in stark contrast to the point-wise approach which requires individual computation for the generation of each pixel. Moreover, our approach exhibits a notable improvement in MAE and MSE, outperforming the point-wise method in a context of a reduced training set and an expanded test set. These enhancements in accuracy and efficiency position the dense estimation approach as a highly effective and scalable alternative for large-scale environmental monitoring.

\paragraph{Application to a New Geographical Region - US West Coast}

An integral part of evaluating the robustness of our approach involved testing the model on a dataset consisting of locations along the US West Coast. This subset of data comprises 91 samples and was not included in the initial training or validation phases, ensuring an unbiased evaluation of the model's robustness. It features unique characteristics not encountered in the European data used for model training, such as dense, rectangular street patterns and expansive desert areas, which are distinctive to the US West Coast \cite{scheibenreif-2022}.

\begin{figure}[h]
    \centering
    \includegraphics[width=\columnwidth]{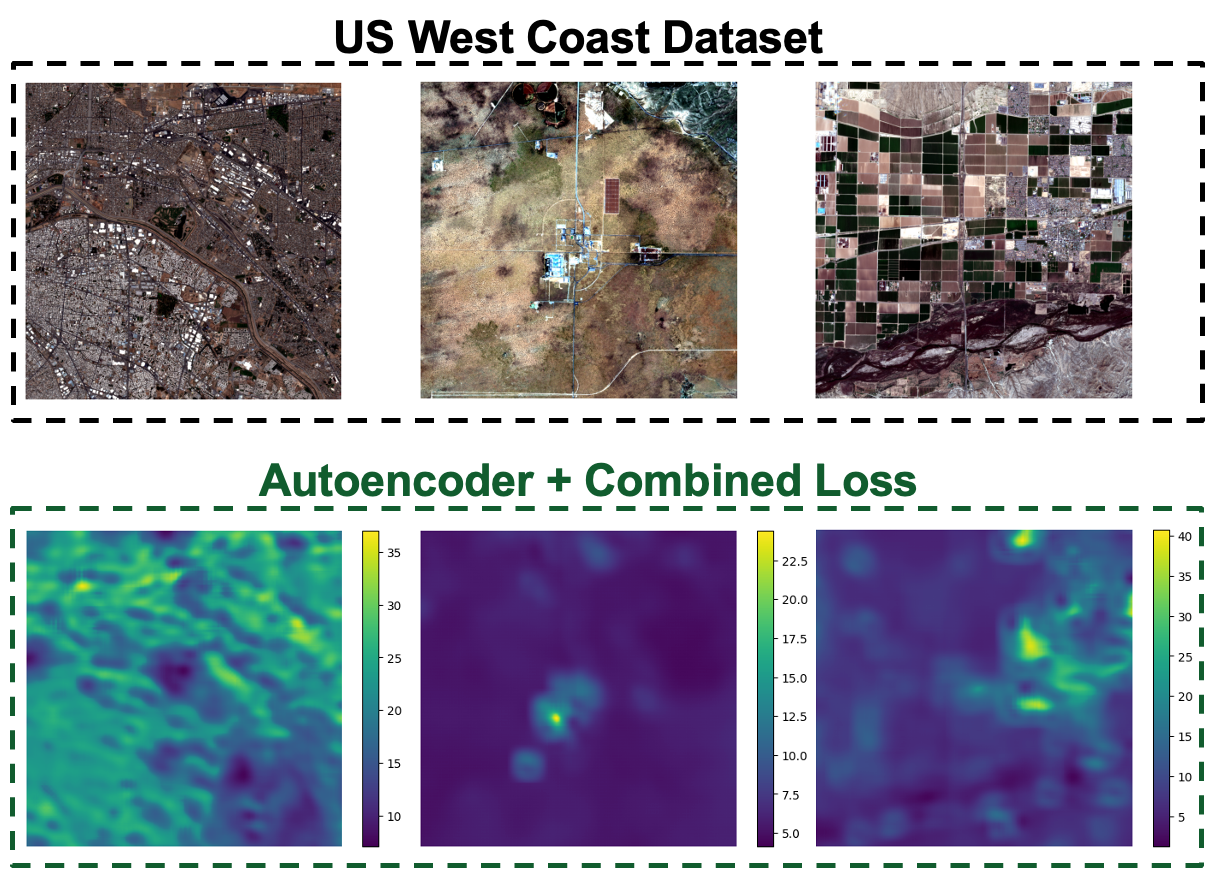}
    \caption{Application of the Autoencoder model with a combined loss on the US West Coast dataset, showcasing the model's capability to adapt and estimate NO\textsubscript{2} levels in diverse geographic settings.}
    \label{fig:us_results}
\end{figure}

\begin{table}[!ht]
  \centering
  \begin{tabular}{@{}l c c c c}
    \toprule
\textbf{Model} & \textbf{Loss} & \textbf{R2 ($\uparrow$)} & \textbf{MSE ($\downarrow$)} & \textbf{MAE ($\downarrow$)} \\
    \midrule
UNet (Ours) & Combined & -1.41 & 140.69 & 10.17 \\ 
UNet (Ours) & NO\textsubscript{2} & -1.42 & 140.74 & 10.09 \\ 
\textbf{AE (Ours)} & Combined & 0.15 & \textbf{62.18} & \textbf{6.19} \\ 
AE (Ours) & NO\textsubscript{2} & -0.02 & 71.43 & 6.69 \\ 
\multicolumn{2}{@{}l}{Point-wise \cite{scheibenreif-2022}} & \textbf{0.28} & 89.92 & 7.86 \\ 
    \bottomrule
  \end{tabular}
  \caption{Evaluation on US West Coast Dataset.}
  \label{tab:us_results}
\end{table}

When evaluated based on the MAE, the dense estimation approach, using the Autoencoder architecture, not only adapted to this new environment but also surpassed the quantitative results of the point-wise approach.

The performance of the dense estimation approach in this new region is indicative of its potential for broader application and underscores the benefits of an adaptable and efficient model for global-scale environmental monitoring.

\paragraph{Impact of Increased Prediction Space}
In a continued effort to optimize our model's performance, we explored the effects of increasing the prediction space on both the quality and efficiency of our NO\textsubscript{2} concentration estimations.


Given the qualitative superiority observed in the Autoencoder architecture's output compared to that of the UNet, the former was selected for this phase of evaluation. Similarly, the Combined Loss (\ref{eq:L}) was employed over the Squared Error Loss (\ref{eq:SEL}) due to its slightly superior quantitative performance, particularly in terms of MAE.

\begin{table}[!ht]
  \centering
  \begin{tabular}{@{}c c c c}
    \toprule
\textbf{Prediction Space} & \textbf{R2 ($\uparrow$)} & \textbf{MSE ($\downarrow$)} & \textbf{MAE ($\downarrow$)}  \\
    \midrule
2 x 2 & 0.49 & 50.21 & \textbf{4.95} \\ 
4 x 4 & 0.49 & 49.11 & 5.04 \\ 
8 x 8 & 0.47 & 50.94 & 4.98 \\ 
16 x 16 & 0.46 & 50.15 & 5.00 \\ 
\textbf{32 x 32} & \textbf{0.50} & \textbf{47.12} & \textbf{4.95} \\ 
64 x 64 & 0.49 & 47.13 & 4.99 \\ 
    \bottomrule
  \end{tabular}
  \caption{Evaluation of varying prediction space dimensions.}
  \label{tab:prediction_space_comparison}
\end{table}

Our investigation revealed that enlarging the prediction space did not result in a noticeable compromise in the model's performance. Both quantitative metrics and qualitative assessments remained consistently similar to those obtained with smaller prediction spaces. This finding suggests that the model preserves its predictive integrity and spatial resolution of NO\textsubscript{2} distribution patterns, even as the scope of prediction broadens.

The computational benefits are increasingly apparent with the expansion of the prediction space. As the area of estimation grows, the model's efficiency improves markedly, affirming the scalability of our approach in processing larger geographical extents with reduced computational demands.



\section{Conclusion}
\label{sec:conclusion}

This study introduced an innovative dense estimation approach using deep learning and remote sensing data for estimating ground-level greenhouse gas concentrations, while also tackling the challenge of sparsely available ground-truth annotations. By developing a uniformly random offset sampling technique, we spread the available ground-truth data evenly across designated areas. This technique enables overcoming biases towards central pixel accuracy and enabling comprehensive estimates across wider regions.

The dense estimation approach was developed in response to the limitations of existing point-wise methods, which, while accurate, are computationally intensive and not feasible for global-scale application. By generating a grid of estimates for a designated area simultaneously, the dense estimation approach significantly reduces the computational demands associated with large-scale environmental monitoring.

Our findings demonstrate that this approach not only provides a substantially improved precision compared to point-wise estimations but also enhances computational efficiency significantly. The approach has shown remarkable results in both qualitative and quantitative assessments, enhancing accuracy and maintaining distribution patterns across various scales of prediction space.

Moreover, the applicability of this method was successfully tested on a new geographical region - the US West Coast. The dataset, featuring unique characteristics not encountered in the European data used for model training, provided an unbiased evaluation of the model’s robustness. The dense estimation approach adapted to this new environment and also surpassed the quantitative results of the point-wise approach in terms of MAE. This underscores the potential of the dense estimation method for broader application and its adaptability to different environmental contexts.

In summary, the dense estimation approach presented in this study offers a viable solution for efficient and scalable global environmental monitoring. It balances the need for detailed, high-resolution data with the practical constraints of computational resources, making it a significant contribution to the field of remote sensing and environmental monitoring. This approach opens new avenues for global estimations and improved environmental policy-making, showcasing the potential of deep learning in addressing critical environmental challenges.
{
    \small
    \bibliographystyle{ieeenat_fullname}
    \bibliography{main}
}

\end{document}